\documentclass[10pt,conference]{IEEEtran}

\PassOptionsToPackage{hyphens}{url}
\usepackage[utf8]{inputenc}
\usepackage[T1]{fontenc}
\usepackage{url}
\usepackage{cite}
\usepackage{graphicx}
\usepackage[caption=false,font=footnotesize]{subfig}
\usepackage{float}
\usepackage{placeins}
\usepackage{algorithm}
\usepackage{algorithmic}
\usepackage{amsmath}
\usepackage{amssymb}
\usepackage{array}
\usepackage{booktabs}
\usepackage{colortbl}
\usepackage{microtype}
\usepackage[hidelinks]{hyperref}

\makeatletter
\newcommand{\tablecaptionof}[1]{\def\@captype{table}\caption{#1}}
\makeatother

\renewcommand{\texttt}[1]{\textrm{#1}}

\title{Janus: An Algorithm--Evaluator Co-Evolution Framework for LLM-Driven Discovery under Expensive Evaluation Budgets}

\author{%
  \IEEEauthorblockN{%
    Ximeng Liu\IEEEauthorrefmark{1}\IEEEauthorrefmark{2}\IEEEauthorrefmark{3},
    Qianlong Wang\IEEEauthorrefmark{3},
    Yingming Mao\IEEEauthorrefmark{4},
    Annan Li\IEEEauthorrefmark{3},
    Yatao Li\IEEEauthorrefmark{5},\\
    Shizhen Zhao\IEEEauthorrefmark{1},
    Jianmin Wu\IEEEauthorrefmark{3},
    Dawei Yin\IEEEauthorrefmark{3},
    Dou Shen\IEEEauthorrefmark{3}}
  \IEEEauthorblockA{%
    \IEEEauthorrefmark{1}Shanghai Jiao Tong University, Shanghai, China\\
    \IEEEauthorrefmark{2}Zhongguancun Academy, Beijing, China\\
    \IEEEauthorrefmark{3}Baidu Inc., Beijing, China\\
    \IEEEauthorrefmark{4}Xi'an Jiaotong University, Xi'an, China\\
    \IEEEauthorrefmark{5}Zhongguancun Institute of Artificial Intelligence, Beijing, China}
}

\hypersetup{
  hidelinks,
  pdftitle={Janus: An Algorithm--Evaluator Co-Evolution Framework for LLM-Driven Discovery under Expensive Evaluation Budgets},
  pdfauthor={Ximeng Liu, Qianlong Wang, Yingming Mao, Annan Li, Yatao Li, Shizhen Zhao, Jianmin Wu, Dawei Yin, Dou Shen}
}

\begin{document}

\maketitle

% !TeX root = ../main.tex
\begin{abstract}
LLM-driven program discovery relies on rapid evaluator feedback, but many
scientific and engineering tasks require high-fidelity simulations, hardware
execution, or physical experiments, making each evaluation expensive. Cheap
surrogate evaluators can reduce this cost, yet fixed surrogates are vulnerable
to search-induced distribution shift and are difficult to fit reliably from
sparse, search-biased labels. We introduce \textbf{Janus}, a framework that
uses LLMs to co-evolve target programs and executable proxy evaluators. To
address label scarcity, Janus leverages domain knowledge encoded in LLMs to
generate task-specific evaluator programs and calibrates them using real
outcomes. To mitigate distribution shift, Janus evolves evaluators alongside
target programs, selects them using a promotion-aligned objective, and
maintains region-conditioned portfolios with online credit updates. Because
proxy predictions remain fallible, Janus uses them only to prioritize
candidates and requires real validation before candidates can enter the
target-program population or update the incumbent. Across five scientific and
engineering design tasks, Janus achieves a larger area under the best-so-far
improvement curve over the real-evaluation budget and higher final performance
than a matched baseline that evolves only target programs. On average, Janus
reaches $99\%$ of the baseline's final improvement with $59.1\%$ fewer real
evaluations. Evolved proxy evaluators also rank promising candidates more
accurately than their seed versions. Together, these results extend
evaluator-guided LLM discovery from tasks with cheap, scalable feedback to
scientific domains where trustworthy evaluation is scarce and expensive.
\end{abstract}

% !TeX root = ../main.tex
\section{Introduction}

Large language models (LLMs) are increasingly being used to discover
high-performing executable programs
\cite{yang2024opro,ye2024reevo,grayeli2024symbolic,liu2025synthesizing}. For
example, systems such as FunSearch and AlphaEvolve embed an LLM within an
iterative discovery loop, in which the model proposes programs, an evaluator
measures their quality, and the resulting feedback guides subsequent
generations
\cite{romeraparedes2024funsearch,novikov2025alphaevolve,cheng2025genesys}.
These systems have achieved remarkable results. Their success relies
on the availability of an evaluator that can assess generated programs and
provide timely feedback to guide the LLM's subsequent proposals.

However, many scientific and engineering tasks lack such timely feedback. In these settings, assessing a single candidate may require high-fidelity simulation, hardware execution, or physical experiments that take hours or even days. Evaluating every LLM-generated proposal is therefore impractical, making real evaluation, rather than candidate generation, the primary bottleneck. This mismatch between abundant candidate generation and scarce real-world feedback limits the applicability of existing LLM-driven search frameworks to scientific and engineering discovery.

\begin{figure*}[t]
\centering
\begin{minipage}[t]{0.60\textwidth}
\centering
\includegraphics[width=\linewidth]{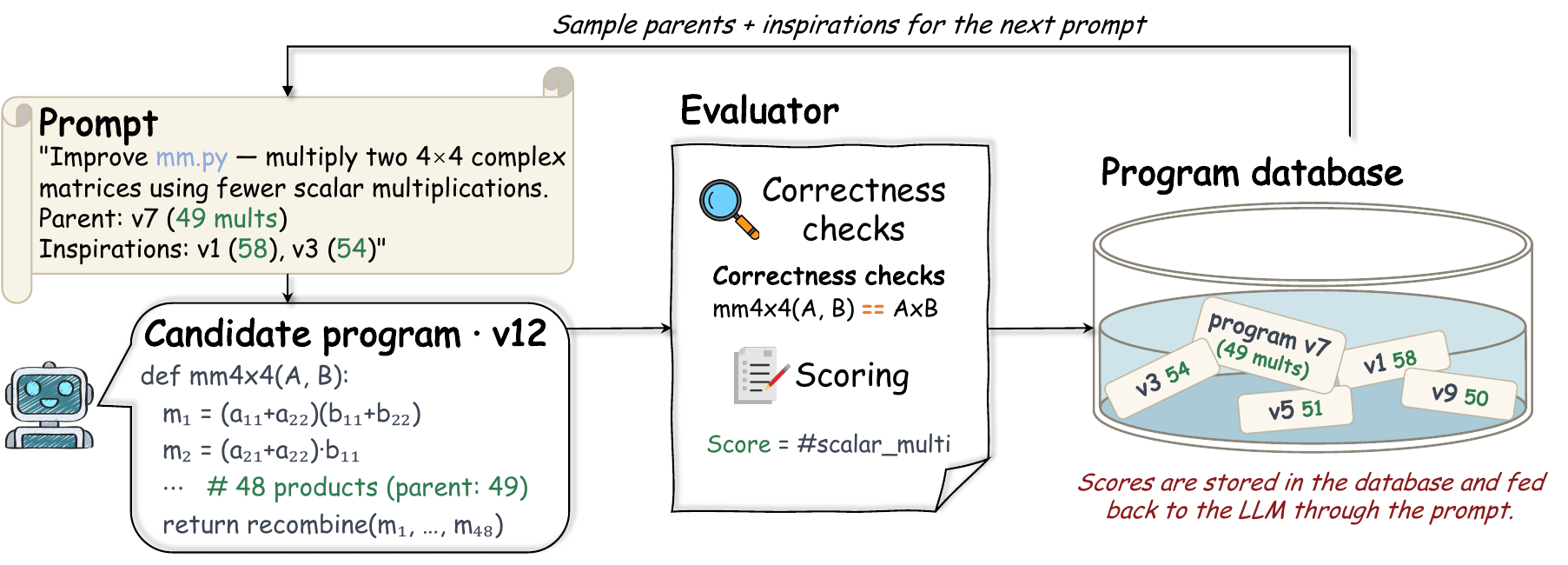}
\caption{AlphaEvolve-style evaluator-guided discovery loop. The LLM proposes executable candidates; automated evaluators score them; the resulting successes and failures update the program database and condition later proposals.}
\label{fig:alphaevolve-loop}
\end{minipage}
\hfill
\begin{minipage}[t]{0.36\textwidth}
\centering
\includegraphics[width=\linewidth]{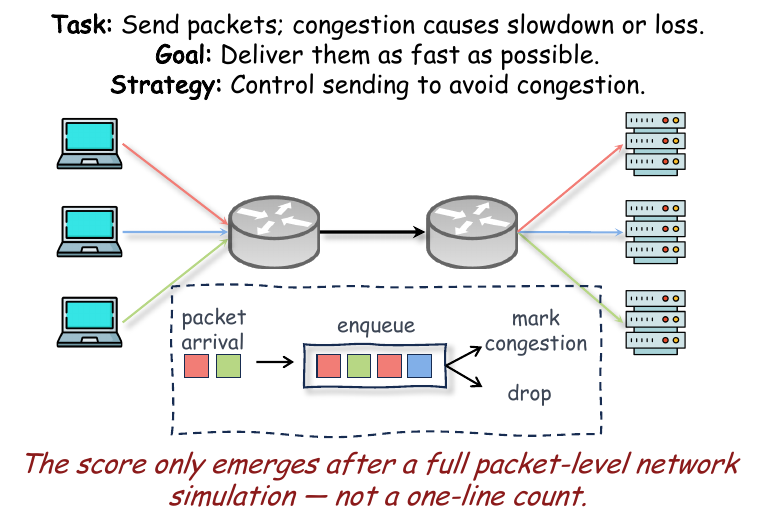}
\caption{Evaluating a congestion-control design requires a suite of packet-level discrete-event simulations.}
\label{fig:ns3-expensive-eval}
\end{minipage}
\end{figure*}

A natural solution is to train a cheap surrogate using a small set of real
evaluations and use it to screen candidates, as commonly done in Bayesian
optimization, surrogate-assisted search, and model-based optimization
\cite{balandat2020botorch,eriksson2019turbo,brookes2019cbas,trabucco2021coms}.
However, this approach faces two major challenges. First,
\textit{distribution shift}: because search is guided by surrogate predictions,
the candidate distribution may move beyond the surrogate's training support,
causing prediction errors to accumulate and ultimately misdirect the search
\cite{yao2024generative}.
Second, \textit{label scarcity}: real evaluations are limited and selected by
the search process, making the available data both sparse and biased. In
scientific problems governed by physical processes, different regions of the
search space may exhibit qualitatively different behaviors as different
physical mechanisms become active. With only sparse labels, a surrogate from a
predefined model family may fail to identify these regime changes and
generalize reliably beyond the observed data.

To address these challenges, we introduce \textit{Janus}, a framework that
co-evolves target programs and executable proxy evaluators under a shared
real-evaluation budget. \textit{To overcome label scarcity}, Janus leverages
domain knowledge encoded in LLMs to generate task-specific executable evaluator
programs with plausible features, equations, and conditional structures, and
calibrates them using limited real outcomes. \textit{To mitigate search-induced
distribution shift}, Janus evolves evaluators alongside target programs,
selects them using a promotion-aligned objective, and maintains
region-conditioned portfolios whose credits are updated online based on real
evaluation results. \textit{Finally, to guard against residual proxy errors},
Janus uses proxy predictions only to screen candidates and allocate the
real-evaluation budget. Candidates remain provisional until they pass real
validation, and the budget is allocated by balancing predicted quality,
uncertainty, and novelty.

We evaluate Janus on five program-design benchmarks: PyBaMM, AQM, FDTD Demux,
Reactor, and Perishable IRP
\cite{rodriguez2024fast,almeida2024desired,wu2024cwdm,braniff2024risk,violi2024perishable}.
Across all five domains, Janus achieves a larger area under the best-so-far
improvement curve over the real-evaluation budget and higher final performance
than a matched baseline that evolves only target programs. On average, Janus
reaches $99\%$ of the final improvement achieved by this baseline using
$59.1\%$ fewer real evaluations. The evolved proxy evaluators also rank
promising candidates more accurately than the initial versions. Together,
these results extend evaluator-guided LLM discovery from tasks with cheap,
scalable feedback to scientific domains where trustworthy evaluation is scarce
and expensive.

\begin{table*}[t]
\centering
\small
\setlength{\tabcolsep}{3.2pt}
\renewcommand{\arraystretch}{1.02}
\begin{tabular}{@{}>{\raggedright\arraybackslash}p{0.16\textwidth}>{\raggedright\arraybackslash}p{0.22\textwidth}>{\raggedright\arraybackslash}p{0.447\textwidth}>{\centering\arraybackslash}p{0.13\textwidth}@{}}
\toprule
\textbf{Problem} & \textbf{Objective / artifact} &
\textbf{Validation procedure} & \textbf{Validation time} \\
\midrule
\multicolumn{4}{l}{\textbf{A. AlphaEvolve: final-result verification}} \\
\addlinespace[2pt]
Fast matrix multiplication & Rank-48 $\langle4,4,4\rangle$ tensor decomposition & Reconstruct the target tensor exactly and score the decomposition rank \cite{novikov2025alphaevolve} & $\approx0.7$ ms \\
\addlinespace[3pt]
11-D kissing number & 593-point integer configuration & Check nonzero vectors and every pairwise separation constraint; score cardinality \cite{novikov2025alphaevolve} & $\approx0.62$ s \\
\midrule
\multicolumn{4}{l}{\textbf{B. Deployment-scale validation in three Janus benchmark domains}} \\
\addlinespace[2pt]
Battery fast charging (physical cells) & Maximize cycle life under a 10-min charge & Cycle a physical cell for 100 cycles (early label) or to failure \cite{severson2019battery,attia2020closedloop} & $\approx4$ d \\
\addlinespace[3pt]
Network congestion control (ns-3 ECN) & Minimize tail flow-completion time & Simulate 5 s in ns-3 on a 384-rack, 6,144-host topology \cite{zhao2023parsimon} & 11--27 h \\
\addlinespace[3pt]
Photonic cavity design (3D FDTD) & Confine light in a compact cavity with minimal leakage & Run one converged 3D MEEP FDTD solve (16 CPU cores, 32 GB) \cite{minkov2014automated} & $\approx10$ h \\
\bottomrule
\end{tabular}
\caption{Evaluation-time comparison between two AlphaEvolve problems and representative scientific problems.}
\label{tab:evaluation-cost}
\end{table*}

% !TeX root = ../main.tex
\section{Motivating Observations}

\subsection{Ground Truth Can Be Expensive}
\label{sec:expensive-ground-truth}

Evaluator-guided discovery is effective when candidate quality is cheap to
verify. Figure~\ref{fig:alphaevolve-loop} sketches this loop for AlphaEvolve.
In fast matrix multiplication, algebraic identities certify correctness and
multiplication count provides the score. This inexpensive feedback enabled
AlphaEvolve to find a $48$-multiplication procedure for two $4\times4$ complex
matrices, improving on the $49$ multiplications required by two levels of
Strassen's algorithm
\cite{strassen1969gaussian,novikov2025alphaevolve}.

Many scientific tasks lack such cheap certificates and instead require
high-fidelity simulation, hardware execution, or physical experiments.
Figure~\ref{fig:ns3-expensive-eval} illustrates this problem in congestion
control: evaluating one configuration requires packet-level ns-3 simulations
across network settings, workloads, and random seeds, and can take hours
\cite{henderson2008ns3,zhao2023parsimon}.
Table~\ref{tab:evaluation-cost} shows the broader contrast: the AlphaEvolve
examples take less than a second to verify, whereas representative scientific
evaluations require hours or days. Evaluating every proposal is therefore
impractical in these settings.

\subsection{Fixed Proxies Are Not Enough}
\label{sec:fixed-proxies}

A natural response is to train a cheap surrogate evaluator once and use it in place of the expensive ground truth.
This idea is well established in Bayesian optimization, surrogate-assisted search, and offline model-based optimization: a learned model approximates an expensive objective, and the optimizer queries this model much more frequently than it queries the real evaluator \cite{snoek2012practical,balandat2020botorch,trabucco2021coms}.
However, a proxy may become inaccurate when evolution moves beyond its training distribution.
It may then mistake a poor candidate for a good one and misguide the subsequent search.

We illustrate this failure in protein engineering on the measured green fluorescent protein (GFP) fitness landscape \cite{sarkisyan2016local}. A proxy trained near the wild-type sequence guides search over the broader sequence space. As selected candidates drift away from the training centroid, proxy scores rise to about 5.6 while measured fluorescence falls to about 2.9 (Figure 3), demonstrating search-induced exploitation of a frozen proxy.

\begin{figure}[t]
\centering
\subfloat[Score misalignment.\label{fig:fixed-proxy-score}]{%
\includegraphics[width=0.485\columnwidth]{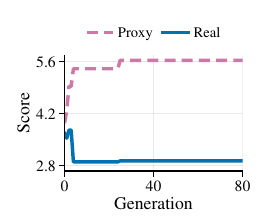}}
\hfill
\subfloat[Feature-space drift.\label{fig:fixed-proxy-support}]{%
\includegraphics[width=0.485\columnwidth]{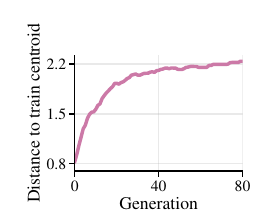}}
\caption{Fixed-surrogate exploitation in GFP design.}
\label{fig:fixed-proxy-drift}
\end{figure}

\subsection{Harnessing LLMs for Adaptive Proxies}
\label{sec:llm-adaptive-proxies}

\S~\ref{sec:fixed-proxies} shows that a proxy must track the distribution induced by search.
Yet refitting an arbitrary proxy online is insufficient: scarce high-fidelity labels leave each refresh with a small, search-biased archive.
Scientific objectives can change qualitatively across different regions of the input space as different physical mechanisms become active.
For example, a network can move abruptly from a low-loss regime to congestion once the offered load exceeds its service and buffering capacity.
A conventional surrogate starts from a predefined model class, such as a GP, random forest, or MLP, and estimates its parameters only from labeled samples.
It must discover such transitions from data alone, which is difficult with a small, search-biased archive.

LLMs offer a way to introduce domain structure before fitting.
We prompt an LLM with a description of the scientific problem and ask it to generate executable proxy functions.
Drawing on its domain knowledge, the LLM can propose features, equations, and conditional branches that encode plausible physical mechanisms.
Real labels are then used to select among the generated functions and calibrate their parameters.
This narrows the search to physically plausible structures, making better use
of scarce labels.

\begin{figure}[t]
\centering
\subfloat[Diagnostic context A.\label{fig:fewshot-incast-diagnostic}]{%
\includegraphics[width=0.49\columnwidth]{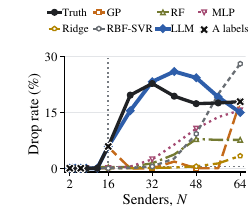}}
\hfill
\subfloat[Knee error (grid steps).\label{fig:fewshot-incast-knee-error}]{%
\includegraphics[width=0.49\columnwidth]{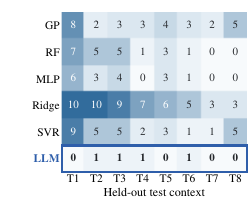}}
\caption{Few-shot TCP-incast knee recovery.}
\label{fig:fewshot-incast-regime}
\end{figure}

To illustrate this advantage, we consider TCP NewReno incast in ns-3.
Given a network setting, the task is to predict how the packet-drop rate changes with the number of concurrent senders and identify the congestion knee, defined as the first tested sender count at which the drop rate reaches $0.5\%$.
Every proxy is trained on the same 12 labeled simulations, with four sender counts measured under each of three training network settings, and is tested on eight unseen settings.
We compare GP, random-forest, MLP, quadratic-ridge, and RBF-SVR proxies with an LLM-generated arithmetic proxy \cite{snoek2012practical,gorissen2009evolutionary}.
Figure~\ref{fig:fewshot-incast-regime}(a) shows that the LLM proxy most closely follows the measured drop-rate curve and captures the transition at $N=16$ in one training setting.
Figure~\ref{fig:fewshot-incast-regime}(b) shows that it also has the lowest mean knee-location error across the eight unseen settings: $0.500$ tested sender-count positions, compared with $2.125$ for the strongest conventional baseline.
The LLM can draw on networking knowledge to encode the capacity-driven transition and post-congestion behavior before sparse labels are used for calibration.

\FloatBarrier

Together, the insights from \S~\ref{sec:fixed-proxies} and \S~\ref{sec:llm-adaptive-proxies} motivate Janus.
To extend evaluator-guided discovery to scientific domains where real evaluation is expensive, Janus uses an LLM to generate and update proxy evaluators online from the task description and accumulated real labels.
Janus therefore co-evolves not only target programs but also the proxy evaluators that guide the search.

% !TeX root = ../main.tex
\section{Janus Design}
\label{sec:janus-design}

\subsection{Overview}
\label{sec:design-overview}

\noindent\textbf{Problem statement.}
We study program discovery for scientific tasks, where candidate programs must be evaluated by an expensive real evaluator.
Let $\mathcal{X}$ denote the program space, and let $f_{\mathrm{real}}(x)$ return a utility $y(x)\in\mathbb{R}$ and a validity label $v(x)\in\{0,1\}$ for each candidate $x\in\mathcal{X}$.
Given a budget of $B$ real-evaluation attempts, the objective is to find the valid program with the highest observed real utility:
\begin{equation}
x_B^{\star}\in
\arg\max_{(x,y,v)\in\mathcal{H}_B:\,v=1} y,
\label{eq:budget-objective}
\end{equation}
where $\mathcal{H}_B$ denotes the real-evaluation outcomes.

\noindent\textbf{Framework components.}
Janus builds on AlphaEvolve's island-based LLM program-search framework
\cite{novikov2025alphaevolve}. Each island $i$ maintains a real-validated
algorithm population $P_{\mathrm{alg}}^i$, while the LLM generates offspring
from selected parents. Janus additionally maintains an evaluator population
$P_{\mathrm{eval}}$, a real-outcome archive $\mathcal{A}$ for fitting and
selecting evaluators, and an island-local pool $\mathcal{C}_i$ of proxy-scored
candidates awaiting real evaluation. An active portfolio $\Pi$, selected from
$P_{\mathrm{eval}}$, screens and prioritizes candidates.

\noindent\textbf{Framework pipeline.}
As shown in Algorithm~\ref{alg:coevolution}, Janus interleaves target proposal,
real evaluation, evaluator evolution, and archive refresh. LLM-generated target
programs are provisionally scored by the active evaluator portfolio; selected
candidates are then real-evaluated to update $\mathcal{H}$, $\mathcal{A}$, and
the valid target population. In parallel, LLM-generated evaluators are fitted
and selected on $\mathcal{A}$, while each archive refresh updates $\Pi$ and
re-scores provisional candidates.

To make this pipeline work, Janus represents proxy evaluators as LLM-generated
programs (\S~\ref{sec:evaluator-program}), selects evaluator portfolios for
target-program clusters using a promotion-aligned meta-objective and updates
their credit from online real outcomes
(\S~\ref{sec:archive-metaevolution}), and prevents proxy errors from
contaminating the target population through exploration-aware, real-validated
promotion (\S~\ref{sec:real-promotion}).

\begin{algorithm}[!t]
\caption{Janus Pipeline}
\label{alg:coevolution}
\small
\begin{algorithmic}[1]
\REQUIRE LLM; $f_{\mathrm{real}}$; budget $B$; promotion period $\tau_p$; evaluator-evolution period $\tau_e$; refresh batch size $b$
\STATE Initialize all state and attempt count $a$
\WHILE{$a<B$}
    \STATE Choose island $i$ with local step $t_i$
    \IF{$t_i\bmod\tau_p=0$}
        \STATE \textit{Real evaluation:} choose $x$ from $\mathcal{C}_i$ or generate it directly
        \STATE Evaluate $x$ with $f_{\mathrm{real}}$; update $\mathcal{H}$, $\mathcal{A}$, and $a$
        \STATE If $v(x)=1$, update $P_{\mathrm{alg}}^i$
    \ELSIF{$t_i\bmod\tau_e=0$ and $|\mathcal{A}|$ is sufficient}
        \STATE \textit{Evaluator evolution:} generate $e$ by LLM mutation from $P_{\mathrm{eval}}$
        \STATE Fit and meta-evaluate $e$ on $\mathcal{A}$; update $P_{\mathrm{eval}}$ and $\Pi$
    \ELSIF{$\Pi\neq\varnothing$}
        \STATE \textit{Target proposal:} generate $x$ by LLM mutation from $P_{\mathrm{alg}}^i$
        \STATE Score $x$ with the region-conditioned subset $\Pi_i(x)\subseteq\Pi$
        \STATE Enqueue $x$ provisionally in $\mathcal{C}_i$
    \ELSE
        \STATE \textit{Cold start:} generate $x$ from $P_{\mathrm{alg}}^i$ and evaluate it with $f_{\mathrm{real}}$
        \STATE Update $\mathcal{H}$, $\mathcal{A}$, $a$, and, if valid, $P_{\mathrm{alg}}^i$
    \ENDIF
    \IF{$b$ new real outcomes have entered $\mathcal{A}$ since the last refresh}
        \STATE \textit{Archive refresh:} refit evaluators, update $\Pi$, and re-score candidates
    \ENDIF
\ENDWHILE
\RETURN $\arg\max_{(x,y,v)\in\mathcal{H}:\,v=1}y$
\end{algorithmic}
\end{algorithm}

\subsection{Evaluator-as-Program}
\label{sec:evaluator-program}

\noindent\textbf{Evaluator representation.}
Rather than fixing a surrogate family in advance, Janus represents each proxy evaluator as an executable program:
\begin{equation}
e=\bigl(g_{\phi},\theta_{\phi}(\mathcal{A})\bigr),
\label{eq:evaluator-representation}
\end{equation}
where $g_{\phi}$ is the program structure generated by an LLM, and $\theta_{\phi}(\mathcal{A})$ contains the values of any adjustable parameters in that program after they have been fitted using the real-outcome archive $\mathcal{A}$.
For example, these parameters may include coefficients in a generated equation, thresholds in conditional branches, or parameters used to calibrate the program's outputs.
The dependence on $\mathcal{A}$ indicates that these values are updated as new real outcomes enter the archive.
Separating the generated structure from its fitted parameters allows Janus to adapt an existing evaluator to new observations without asking the LLM to regenerate its code.
If a generated program has no adjustable parameters, $\theta_{\phi}(\mathcal{A})$ is empty.
The framework requires only a common interface for producing a candidate-selection score; within it, the LLM may construct features, equations, conditional branches, low-fidelity computations, or learned corrections.

\noindent\textbf{LLM-based evaluator generation.}
At each evaluator-evolution event, Janus selects a parent and structurally diverse inspirations from the clustered evaluator population $P_{\mathrm{eval}}$.
The LLM receives the scientific task and real-evaluation specifications, the selected evaluator programs, and archive-derived diagnostics and judge feedback.
It then outputs an executable Python evaluator implementing the required interface.
The framework checks the generated artifact, executes it under a timeout, fits $\theta_{\phi}(\mathcal{A})$, and evaluates it on $\mathcal{A}$ using the promotion-aligned objective introduced in \S~\ref{sec:archive-metaevolution}.
Evaluators that pass these checks can enter $P_{\mathrm{eval}}$, and their observed ranking errors and judge feedback are included in subsequent LLM prompts.
Thus, the LLM proposes the proxy structure, while real outcomes calibrate its parameters and determine whether it is retained.

\subsection{Evaluator Selection and Adaptation}
\label{sec:archive-metaevolution}

Janus first maintains the real-outcome data used to fit proxy evaluators, then fits and selects evaluators according to their ability to identify promising candidates, and finally adjusts their use based on online performance.

\noindent\textbf{Real-outcome archive.}
The archive $\mathcal{A}$ is bounded, deduplicated, and balanced rather than maintained as a first-in, first-out history.
It retains high-utility valid programs to preserve the real frontier, recent programs to reflect the current search distribution, valid--invalid boundary cases to expose feasibility errors, and structurally diverse programs to avoid duplicate-heavy evaluator assessment.
After a batch of new real labels arrives, Janus advances the archive revision, refits and re-evaluates the leading evaluators, reselects $\Pi$, and re-scores stale provisional candidates.

\noindent\textbf{Evaluator fitting and selection.}
Ultimately, the proxy evaluator serves to help search discover better programs.
Accurately identifying the highest-utility candidates therefore matters more than minimizing average prediction error across the archive.
Janus accordingly scores each fitted evaluator by how well its top ranking recovers the best programs in $\mathcal{A}$:
\begin{equation}
F_{\mathrm{meta}}(e)=
\lambda_R\bigl(1-R_k(e)\bigr)
+\lambda_N N_k(e)
+\lambda_{\rho}\rho(e),
\label{eq:meta-fitness}
\end{equation}
where $R_k(e)$ is the normalized regret, indicating whether the evaluator's predicted top-$k$ contains a near-optimal program, $N_k(e)$ is the normalized discounted cumulative gain (NDCG), indicating how closely the predicted top-$k$ matches the true top-$k$ ranking, and $\rho(e)$ measures global ranking consistency across the archive.
The first two terms receive most of the weight because they directly measure the quality of candidates sent to real evaluation, while the correlation term provides a weaker global signal.

\noindent\textbf{Portfolio and online credit.}
Different evaluators may be reliable in different regions of the evolving program distribution.
For each island, Janus treats the clusters in the current target-program population as regions and represents each region by the centroid of its programs' code embeddings.
A new candidate $x$ is assigned to the nearest region according to cosine distance:
\begin{equation}
r(x)=\arg\min_r d_{\mathrm{cos}}\bigl(z(x),\mu_r\bigr),
\label{eq:region-assignment}
\end{equation}
where $r$ ranges over the target-program regions on island $i$, $z(x)$ is the candidate's code embedding, $\mu_r$ is the centroid of region $r$, and $d_{\mathrm{cos}}$ denotes cosine distance.
After determining $r(x)$, Janus selects the $K$ evaluators in the active pool $\Pi$ with the highest current credit in that region, forming the region-specific portfolio $\Pi_i(x)$ that jointly scores $x$.

The meta-evaluation scores used to initialize these credits are refreshed only after a new batch of real labels is incorporated into $\mathcal{A}$.
Between batch refreshes, whenever a promoted candidate receives a real outcome, Janus immediately updates the credit of every evaluator in $\Pi_i(x)$ that supported its promotion:
\begin{equation}
F_{e,r}=(1-\alpha_{e,r})F_{\mathrm{meta}}(e)+\alpha_{e,r}g_{e,r},
\label{eq:region-online-credit}
\end{equation}
Here, $e\in\Pi$ indexes an active evaluator, $r$ indexes an island-local target-program region, and $F_{e,r}$ is the current credit of evaluator $e$ in region $r$.
$F_{\mathrm{meta}}(e)$ is its batch-level score from Eq.~\eqref{eq:meta-fitness}, $n_{e,r}$ is the number of candidates it helped promote in $r$ that have received real evaluations, and $\kappa>0$ controls how quickly this evidence overrides the batch-level score.
Thus, $\alpha_{e,r}=n_{e,r}/(n_{e,r}+\kappa)$ weights the regional evidence, while $g_{e,r}$ combines the fraction of these candidates that are valid and improve the incumbent with their normalized mean improvement.
Valid improvements therefore increase an evaluator's regional credit, whereas invalid or non-improving recommendations reduce its weight for subsequent candidates in the same region.

\noindent\textbf{Portfolio prediction.}
For a candidate $x$, Janus obtains three quantities from each evaluator $e\in\Pi_i(x)$: a normalized score $\hat{s}_e(x)\in[0,1]$, a predicted validity probability $p_{v,e}(x)$, and an uncertainty estimate $\sigma_e$.
The evaluator produces the first two, while Janus estimates $\sigma_e$ as the standard deviation of its normalized score errors on the real-valid archive entries.
Janus normalizes the region-specific credits $F_{e,r(x)}$ into weights and uses the weighted averages of $\hat{s}_e(x)$ and $p_{v,e}(x)$ as the portfolio score $\bar{s}(x)$ and validity probability $\bar{p}_v(x)$.
The portfolio uncertainty $\bar{\sigma}(x)$ combines the weighted evaluator uncertainties with the weighted standard deviation of their scores on $x$, thereby capturing both historical prediction error and disagreement within the portfolio.

\begin{figure*}[!t]
\centering
\newlength{\caserowheight}
\setlength{\caserowheight}{0.24\textwidth}
\begin{minipage}[t][\caserowheight][c]{0.315\textwidth}
\vspace{0pt}
\centering
\scriptsize
\setlength{\tabcolsep}{1.2pt}
\renewcommand{\arraystretch}{1.18}
\begin{tabular}{@{}>{\raggedright\arraybackslash}p{0.24\linewidth}
                    >{\raggedright\arraybackslash}p{0.23\linewidth}
                    >{\raggedright\arraybackslash}p{0.48\linewidth}@{}}
\toprule
\textbf{Domain} & \textbf{Evaluator} & \textbf{Design goal} \\
\midrule
\textbf{Battery} & PyBaMM & Fast charge; cycle life \\
\textbf{Network} & ns-3 AQM & Throughput; delay/loss \\
\textbf{FDTD Demux} & 2D FDTD & Low-crosstalk routing \\
\textbf{Reactor} & Stiff ODE & Yield; thermal safety \\
\textbf{Perishable IRP} & Routing sim. & Low shortage/waste \\
\bottomrule
\end{tabular}
\tablecaptionof{Cases and real evaluators.}
\label{tab:cases}
\end{minipage}
\hfill
\begin{minipage}[t][\caserowheight][t]{0.315\textwidth}
\vspace{0pt}
\centering
\includegraphics[width=\linewidth]{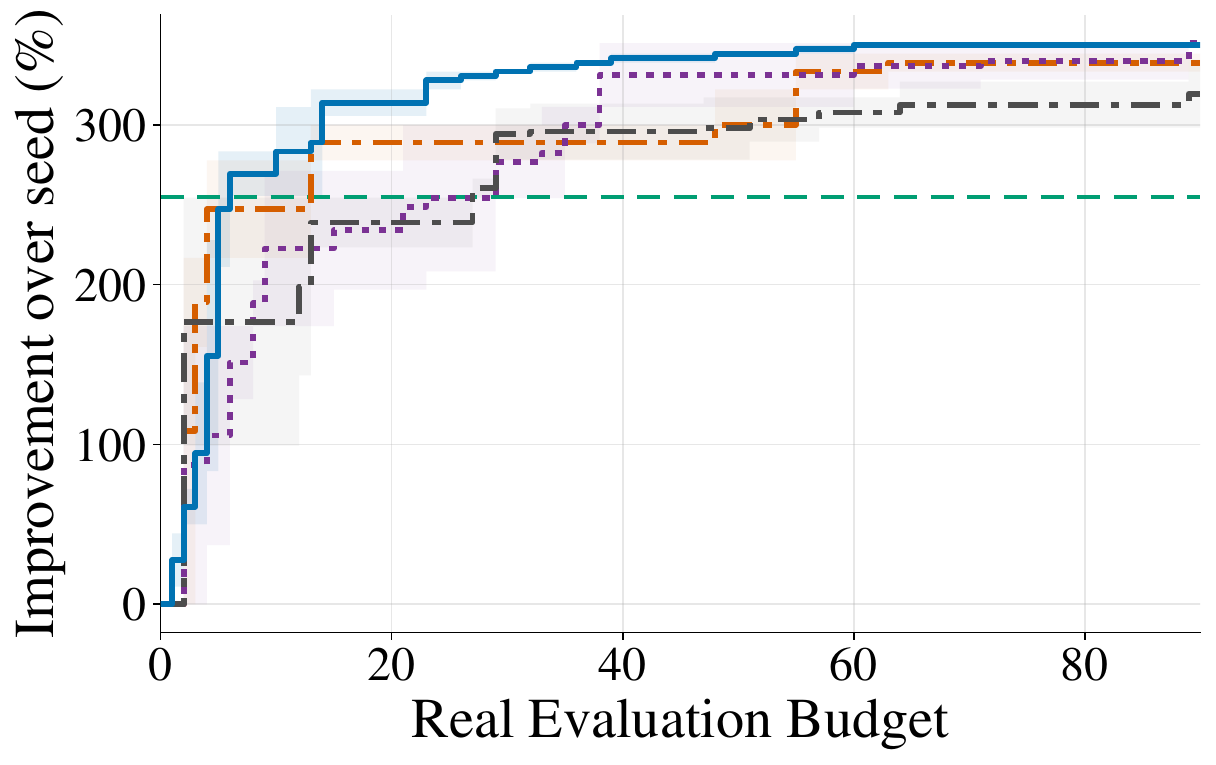}\\
{\footnotesize (a) PyBaMM}
\end{minipage}
\hfill
\begin{minipage}[t][\caserowheight][t]{0.315\textwidth}
\vspace{0pt}
\centering
\includegraphics[width=\linewidth]{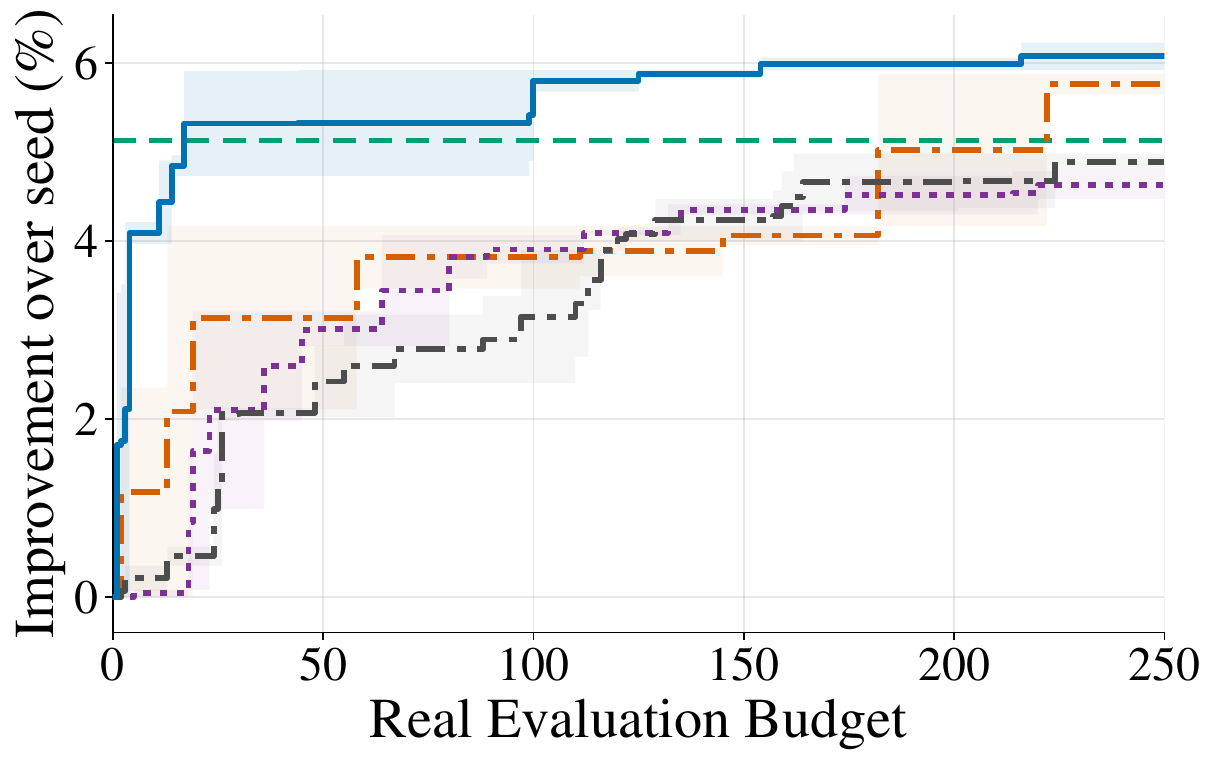}\\
{\footnotesize (b) AQM}
\end{minipage}
\par\vspace{1.5mm}
\begin{minipage}[t]{0.315\textwidth}
\vspace{0pt}
\centering
\includegraphics[width=\linewidth]{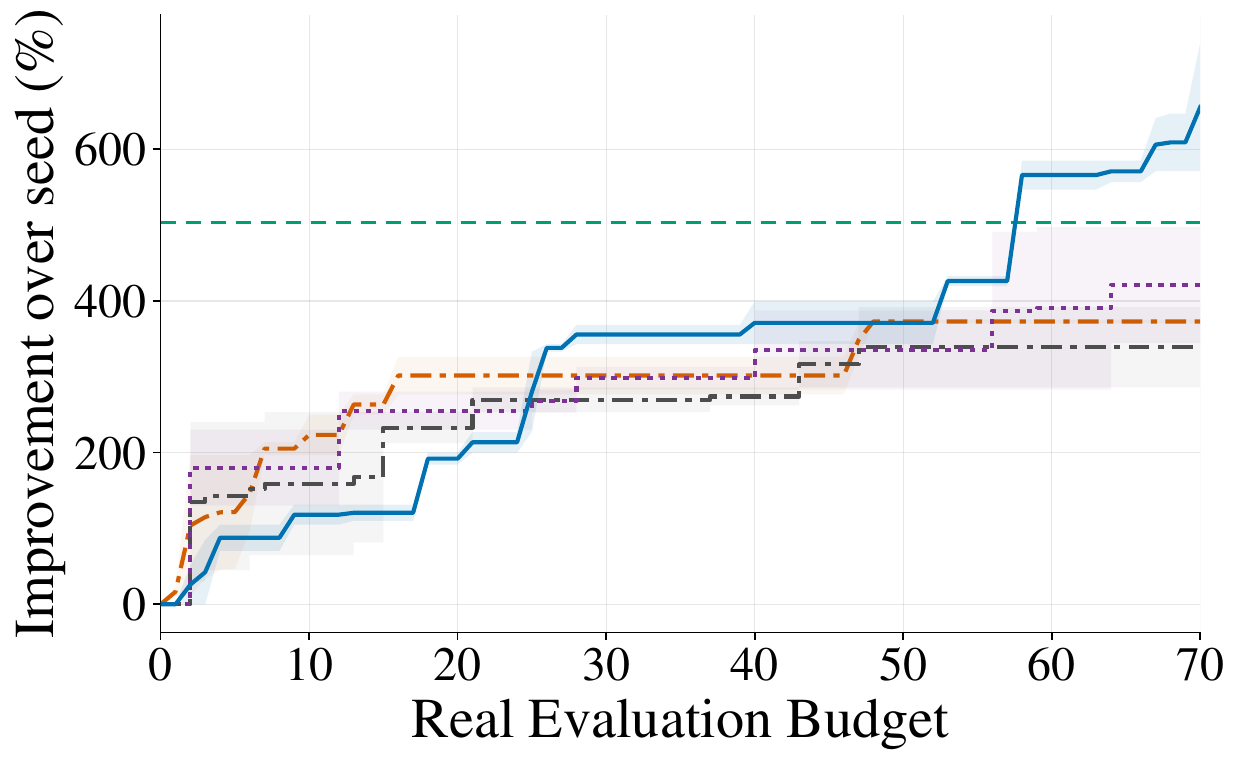}\\
{\footnotesize (c) FDTD Demux}
\end{minipage}
\hfill
\begin{minipage}[t]{0.315\textwidth}
\vspace{0pt}
\centering
\includegraphics[width=\linewidth]{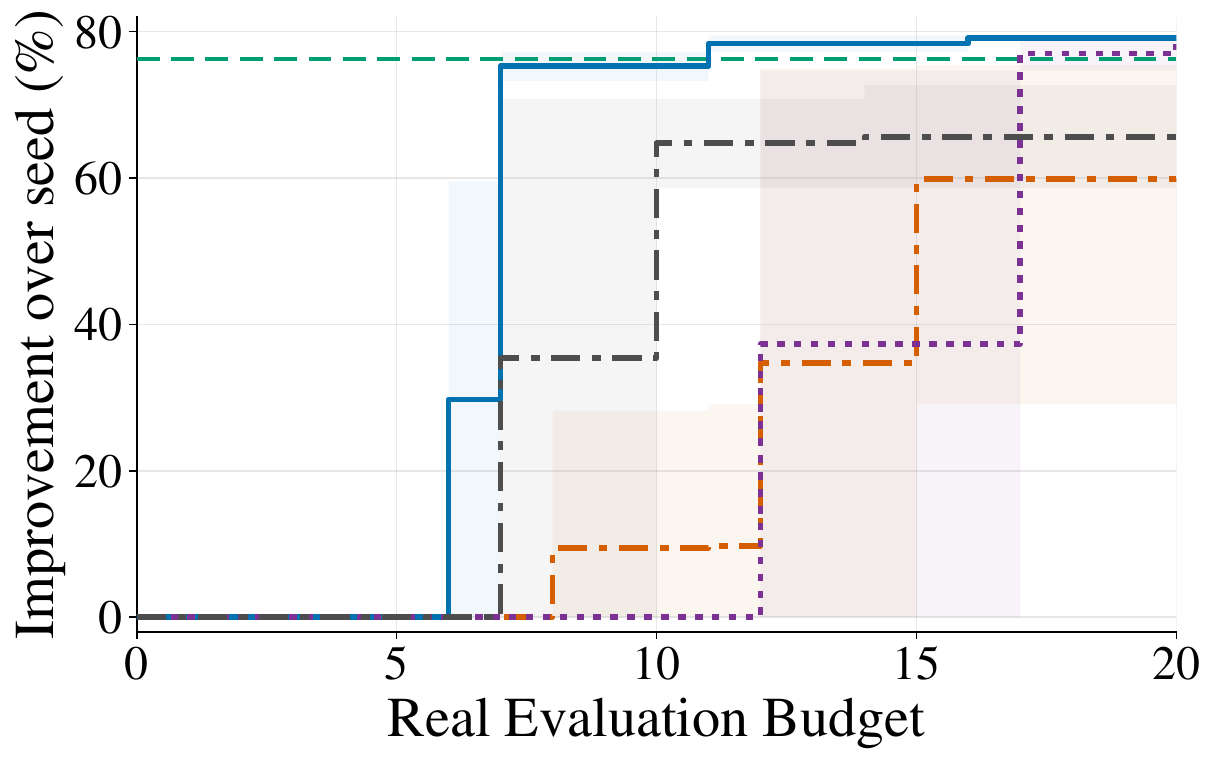}\\
{\footnotesize (d) Reactor}
\end{minipage}
\hfill
\begin{minipage}[t]{0.315\textwidth}
\vspace{0pt}
\centering
\includegraphics[width=\linewidth]{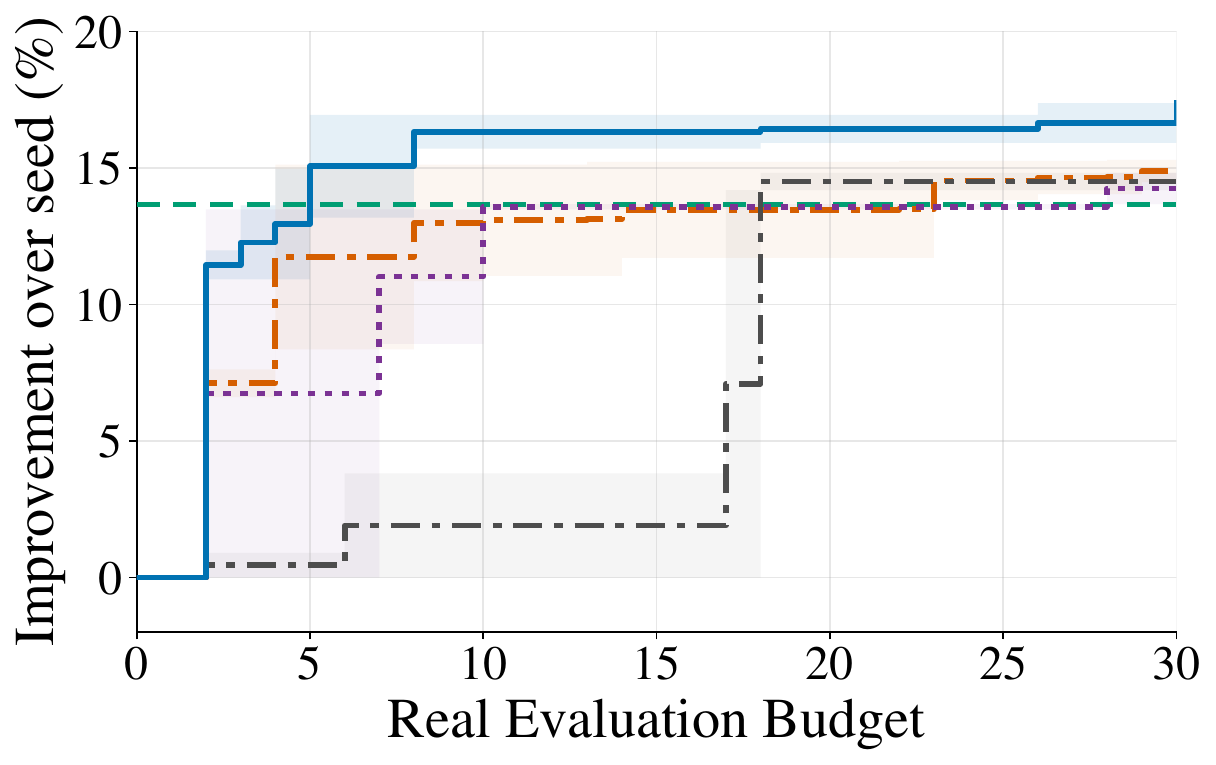}\\
{\footnotesize (e) Perishable IRP}
\end{minipage}
\par\vspace{0.2mm}
\includegraphics[width=0.70\textwidth]{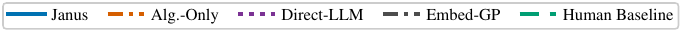}
\vspace{-1.0mm}
\caption{Results across five program-design domains under matched real-evaluation budgets.}
\label{fig:case-results}
\end{figure*}

\subsection{Real-Anchored Promotion}
\label{sec:real-promotion}

\noindent\textbf{Provisional screening.}
Even after selection and online adaptation, proxy evaluators remain fallible under distribution shift, calibration error, or incorrect output scales.
Janus therefore uses their predictions only to allocate the real-evaluation budget.
Between real-evaluation events, the LLM generates offspring from real-validated parents, and the portfolio defined in \S~\ref{sec:archive-metaevolution} produces $(\bar{s}(x),\bar{p}_v(x),\bar{\sigma}(x))$ for each offspring.
Candidates with usable predictions are stored in a bounded and deduplicated pool $\mathcal{C}_i$ and are rescored when the archive or evaluator portfolio changes.

\noindent\textbf{Budget allocation and validation.}
At each promotion event, Janus ranks candidates that pass basic predicted-validity and quality gates using
\begin{equation}
\alpha(x)=w_q\bar{p}_v(x)\bar{s}(x)
+w_u\bar{p}_v(x)\bar{\sigma}(x)+w_n\eta(x),
\label{eq:acquisition}
\end{equation}
where $\alpha(x)$ is the acquisition score; $\bar{s}(x)$, $\bar{p}_v(x)$, and $\bar{\sigma}(x)$ are the portfolio's predicted quality, validity probability, and uncertainty; and $w_q,w_u,w_n\geq0$ are their respective weights.
Here, $\eta(x)$ measures embedding-region novelty based on the cosine distance from the code embedding of $x$ to its nearest target-region centroid; candidates farther from existing regions are considered more novel.
The first term drives exploitation by favoring candidates expected to be both valid and high-quality.
The second explores uncertain but plausibly valid candidates; multiplying both terms by $\bar{p}_v(x)$ avoids spending budget on candidates predicted to be invalid.
The final term explores underrepresented regions independently of the current validity prediction.
After promotion, Janus discards the proxy predictions and calls $f_{\mathrm{real}}$; only a real-valid outcome may enter $P_{\mathrm{alg}}^i$ or update the incumbent.
Every outcome updates $\mathcal{A}$ and the online credit of its recommending evaluators.
Because acquisition alone could suppress candidates misjudged by the entire portfolio, Janus also periodically sends unranked LLM offspring directly to $f_{\mathrm{real}}$.

% !TeX root = ../main.tex
\section{Experiments}

\subsection{Experimental Setup}
\label{sec:experimental-setup}

\paragraph{Benchmarks.}
We evaluate five program-design benchmarks (Table~\ref{tab:cases}):
PyBaMM (battery fast charging) \cite{rodriguez2024fast},
AQM (ns-3 active queue management) \cite{almeida2024desired}, FDTD Demux
(two-dimensional FDTD wavelength demultiplexing) \cite{wu2024cwdm}, Reactor
(constrained exothermic reactor control) \cite{braniff2024risk}, and
Perishable IRP (stochastic perishable inventory routing)
\cite{violi2024perishable}. Appendix~\ref{sec:benchmark-details}
provides objectives, simulation settings, and invalidity checks.

\noindent\textbf{Comparison baselines.}
We compare Janus with
(1) \textit{Alg.-Only}, a matched algorithm-search baseline that uses the same
task-program seed and real-evaluation budget but does not evolve evaluator
programs, following the AlphaEvolve-style mechanism of evolving target programs
against a fixed evaluator \cite{novikov2025alphaevolve};
(2) \textit{Direct-LLM}, which scores each candidate by prompting the LLM with
the labeled history, following the regression-only LAEA setting
\cite{hao2024llmsurrogate};
(3) \textit{Embed-GP}, which fits a Gaussian-process surrogate on fixed code
embeddings of the real-evaluated candidates and uses its predictions to select
candidates for real evaluation \cite{snoek2012practical}; and
(4) \textit{Human Baselines}: CLO, D-RED, CMT, OCP, and an order-up-to
heuristic for PyBaMM, AQM, FDTD Demux, Reactor, and Perishable IRP,
respectively
\cite{attia2020closedloop,floyd1993red,wu2024cwdm,braniff2024risk,
onggo2019agrifood,violi2024perishable}.

\noindent\textbf{Metrics.}
We report (1) AUBC$_{\Delta}$, the average best-so-far improvement over the
real-evaluation budget; (2) Best$_{\Delta}$@$B$, the terminal improvement; and
(3) Calls@99\%, the real-evaluator calls needed to reach $99\%$ of the
Alg.-Only terminal improvement. For evaluator adaptation, NDCG@3 measures
top-three ranking quality, while normalized best regret@3 measures whether the
predicted top three recover a near-optimal candidate.

\begin{table*}[!t]
\centering
\small
\setlength{\tabcolsep}{2.2pt}
\renewcommand{\arraystretch}{1.12}
\begin{tabular}{@{}>{\centering\arraybackslash}p{0.035\textwidth}
                    >{\raggedright\arraybackslash}p{0.20\textwidth}
                    *{3}{>{\centering\arraybackslash\footnotesize}p{0.173\textwidth}}
                    >{\centering\arraybackslash\footnotesize}p{0.185\textwidth}@{}}
\toprule
& & \multicolumn{2}{c}{\textbf{AQM ($B=200$)}} &
\multicolumn{2}{c}{\textbf{FDTD Demux ($B=70$)}} \\
\cmidrule(lr){3-4}\cmidrule(lr){5-6}
\textbf{ID} & \textbf{Change} &
\shortstack{\textbf{AUBC$_{\Delta}$}\\\textbf{(\%)} $\uparrow$} &
\shortstack{\textbf{Best$_{\Delta}$@$B$}\\\textbf{(\%)} $\uparrow$} &
\shortstack{\textbf{AUBC$_{\Delta}$}\\\textbf{(\%)} $\uparrow$} &
\shortstack{\textbf{Best$_{\Delta}$@$B$}\\\textbf{(\%)} $\uparrow$} \\
\midrule
\textbf{F} & \textbf{Full Janus} &
\textbf{$5.5 \pm 0.4$} & \textbf{$6.0 \pm 0.1$} & \textbf{$316.3 \pm 6.9$} & \textbf{$655.7 \pm 119.5$} \\
\midrule
\textbf{S} & No evaluator evolution &
\mbox{$4.1\!\pm\!1.0\;(\downarrow 25.5\%)$} &
\mbox{$5.2\!\pm\!0.7\;(\downarrow 13.3\%)$} &
\mbox{$300.4\!\pm\!45.0\;(\downarrow 5.0\%)$} &
\mbox{$429.9\!\pm\!77.4\;(\downarrow 34.4\%)$} \\
\textbf{A1} & One evaluator only &
\mbox{$4.3\!\pm\!1.6\;(\downarrow 21.8\%)$} &
\mbox{$5.1\!\pm\!1.2\;(\downarrow 15.0\%)$} &
\mbox{$243.9\!\pm\!15.5\;(\downarrow 22.9\%)$} &
\mbox{$340.7\!\pm\!61.7\;(\downarrow 48.0\%)$} \\
\textbf{M0} & Global Spearman objective &
\mbox{$4.8\!\pm\!0.9\;(\downarrow 12.7\%)$} &
\mbox{$5.8\!\pm\!0.1\;(\downarrow 3.3\%)$} &
\mbox{$252.0\!\pm\!31.0\;(\downarrow 20.3\%)$} &
\mbox{$404.7\!\pm\!6.7\;(\downarrow 38.3\%)$} \\
\textbf{O0} & No online credit updates &
\mbox{$4.4\!\pm\!0.2\;(\downarrow 20.0\%)$} &
\mbox{$4.9\!\pm\!0.1\;(\downarrow 18.3\%)$} &
\mbox{$245.1\!\pm\!21.6\;(\downarrow 22.5\%)$} &
\mbox{$382.5\!\pm\!101.5\;(\downarrow 41.7\%)$} \\
\bottomrule
\end{tabular}
\caption{Component ablations on AQM and FDTD Demux.}
\label{tab:component-ablations}
\end{table*}

\subsection{Main Results}

For each method-domain pair, we run three random seeds. Each curve shows mean
best-so-far improvement, with shading spanning the minimum and maximum across
runs.

\noindent\textbf{Consistent budget efficiency.}
Janus achieves both a larger AUBC$_{\Delta}$ and a higher terminal mean than
Alg.-Only in all five domains. The AUBC$_{\Delta}$ gap in budget-averaged
improvement ranges from $1.6$
percentage points on AQM to $45.1$ points on Reactor, indicating that Janus
performs better throughout the search, not only at the endpoint. Janus also
finishes above the strongest human-designed reference shown in every panel, whereas Alg.-Only
remains below the CMT reference on FDTD Demux and the OCP reference on Reactor.

\noindent\textbf{Calls to the Alg.-Only endpoint.}
To compare sample efficiency, we set the target in each domain to $99\%$ of
the final improvement achieved by Alg.-Only and count how many real-evaluator
calls each method needs to reach it. Janus needs $32$ versus $82$
real-evaluator calls on PyBaMM, $100$ versus
$222$ on AQM, $40$ versus $48$ on FDTD Demux, $3$ versus $15$ on Reactor, and
$5$ versus $29$ on Perishable IRP. Thus, Janus reaches the target earlier in
all five domains, reducing the required real-evaluator calls by
$16.7\%$--$82.8\%$ across cases, with a mean reduction of $59.1\%$.

\noindent\textbf{Alternative proxy baselines.}
\textbf{(1) Embed-GP.}
Embed-GP assumes that distances between fixed code embeddings reflect real
utility, although semantic similarity may not preserve the task-specific
numerical or physical behaviors governing performance. It underperforms
Alg.-Only in both AUBC$_{\Delta}$ and final improvement in four of the five
domains: PyBaMM, AQM, FDTD Demux, and Perishable IRP.
\textbf{(2) Direct-LLM.}
Direct-LLM scores each candidate using a growing, heterogeneous history of up
to 50 evaluated programs. Such long prompts may dilute relevant evidence, and
the method does not explicitly model uncertainty or distribution shift. It
therefore achieves lower AUBC$_{\Delta}$ than Janus in all five domains,
particularly on AQM and FDTD Demux, while consuming on average
$8.1\times$ as many LLM tokens per completed real-evaluator call as Janus
(see Appendix~\ref{sec:appendix-token-cost}).

\iffalse
\begin{figure}[H]
\centering
\includegraphics[width=\linewidth]{Figures/pybamm_result.pdf}
\caption{PyBaMM case study result comparing the evaluation band against the real trajectory.}
\label{fig:pybamm-result}
\end{figure}
\fi

\subsection{Evolving Better Evaluators}
\label{sec:evaluator-adaptation}

\begin{figure}[H]
\centering
\includegraphics[width=0.94\linewidth]{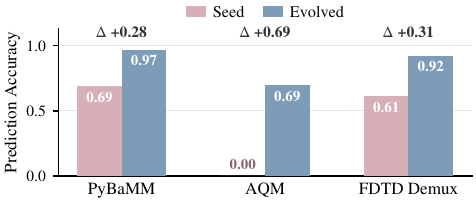}
\caption{Ranking-accuracy gains from evaluator evolution.}
\label{fig:seed-evolution-accuracy}
\end{figure}

\noindent\textbf{Ranking Gains from Evaluator Evolution.}
We test whether evaluator evolution improves the ranking decisions used to
select candidates for real evaluation. For each run, we refit the seed
evaluator and the highest-weight evolved evaluator on the same checkpoint
archive, then compare their NDCG@3 on the same 40 later candidates excluded
from fitting. Figure~\ref{fig:seed-evolution-accuracy} shows that NDCG@3 increases from
$0.688$ to $0.970$ on PyBaMM,
from $0.000$ to $0.694$ on AQM, and from $0.612$ to $0.923$ on FDTD Demux,
giving gains of $0.282$, $0.694$, and $0.311$, respectively.
Appendix~\ref{sec:appendix-evaluator-cases} examines the resulting program changes.
For example, the evolved PyBaMM evaluator adds peak temperature, SEI growth,
voltage headroom, and high-SOC duration, better reflecting the constraints
enforced by the real evaluator.

\noindent\textbf{Zero-shot reuse under objective changes.}
When the objective changes modestly, an existing evaluator can be reused with
a simple adjustment, without new labels or refitting. For example, on the
reweighted FDTD Demux objective, the previously evolved evaluator continues to
rank candidates accurately, achieving an NDCG@3 of 0.959 (see
Appendix~\ref{sec:appendix-zero-shot-reuse}).

% Main-text ablation layout frozen for drafting on 2026-07-18.
% F/S/A1/M0/O0 appear in the main table; R5/R1/SF are reserved for
% finer-grained appendix diagnostics. AQM pilot cells were filled after all
% five arms completed in both domains.
\subsection{Ablation Study}
\label{sec:component-ablations}

We ablate evaluator evolution (S), replace the evaluator portfolio with one
evaluator (A1), replace the promotion-aligned meta-objective with global rank
correlation (M0), and disable online credit updates (O0) on AQM and
FDTD Demux. Table~\ref{tab:component-ablations} reports AUBC$_{\Delta}$ and terminal
improvement over the shared seed under matched real-evaluation budgets, with
full Janus (F) as the reference.
\FloatBarrier

All four ablations reduce both metrics in both domains, showing that the four
components address complementary errors. On AQM, burst traffic distinguishes
controllers with similar average utilization; evolved evaluators capture these
differences through probe-response and burst-specific signals, while limited
headroom makes false-positive promotions especially costly. This explains
AQM's sensitivity to evaluator evolution and online credit updates.
In FDTD Demux, coarse-to-fine simulation error varies across layout
morphologies, favoring a region-weighted portfolio over a single evaluator.
Additional experiments in Appendix~\ref{sec:appendix-llm-backbone} examine the
effect of the LLM backbone. Across the tested backbones, Janus consistently achieves a higher
terminal improvement.

% !TeX root = ../main.tex
\section{Related Work}

\noindent\textbf{LLM-driven algorithm discovery.}
FunSearch and AlphaEvolve place LLM generation inside evaluator-guided
population search, while Evolution of Heuristics and MCTS-AHD improve heuristic
generation and search organization
\cite{romeraparedes2024funsearch,novikov2025alphaevolve,liu2024eoh,zheng2025mctsahd}.
These methods evolve target programs while assuming that the task evaluator is
externally specified. Janus instead makes the executable proxy evaluator part
of the search.

\noindent\textbf{Surrogate-assisted optimization.}
Bayesian optimization and surrogate-assisted evolutionary algorithms reduce
expensive objective calls by fitting cheap response models and selecting which
candidates receive real evaluations
\cite{snoek2012practical,balandat2020botorch,eriksson2019turbo,brookes2019cbas}.
Multi-fidelity methods additionally exploit approximations at different costs
or resolutions \cite{kandasamy2017multifidelity}. Even adaptive approaches
generally fit, evolve, or select models within predefined families or component
libraries \cite{ren2019sacc,gorissen2009evolutionary,akhauri2022eznas,kang2025apd}.
Janus instead uses an LLM to generate task-aware executable proxy structures
from domain descriptions and real outcomes.

\noindent\textbf{LLM-assisted proxies and co-evolution.}
LAEA uses an LLM directly to predict the quality of numerical candidates, while
CoE-SAEA and LLM-SAEA select predefined surrogate models and
candidate-selection rules
\cite{hao2024llmsurrogate,xie2025coesaea,xie2025llmsaea}.
LAEA's numerical reproduction pipeline cannot be transferred unchanged to
free-form programs; our Direct-LLM (LAEA-style) diagnostic retains its core
mechanism of predicting a new candidate directly from historical real
outcomes.
These methods do not generate task-specific executable proxies that evolve with
the search distribution. Related methods adapt rewards, opponents, instances,
or evaluators rather than approximate a fixed, expensive scientific objective
\cite{ma2024eureka,li2025lares,li2026famou,ke2026asro,zhu2026score}.
Janus instead co-evolves executable proxies while real outcomes remain the
source of target fitness.

% !TeX root = ../main.tex
\section{Conclusion}

We introduced Janus, which uses LLMs to co-evolve target programs and
executable proxy evaluators under limited real-evaluation budgets. Janus
calibrates proxy evaluators with real labels, adapts region-conditioned
portfolios as search evolves, and allocates evaluations by balancing
exploitation and exploration. Across five domains, Janus achieves higher
AUBC$_{\Delta}$ and terminal performance than a matched baseline that evolves
only target programs, while reaching the baseline's final improvement
with $59.1\%$ fewer real evaluations.

\bibliographystyle{IEEEtran}
\bibliography{references}

\appendices
\begingroup
\footnotesize
% !TeX root = ../main.tex
\section{Benchmark Details}
\label{sec:benchmark-details}

Each benchmark asks the search procedure to produce an executable program for
a domain-specific design or control problem. The following paragraphs define
the problem, the program output, and the real-evaluation procedure used to
score that output.

\paragraph{PyBaMM battery fast charging.}
The problem is to design a multistage constant-current protocol that reaches
the target state of charge quickly while limiting degradation and satisfying
electrochemical and thermal constraints
\cite{attia2020closedloop,rodriguez2024fast}. The target program implements
\emph{solve(problem)} and returns charge stages, each specifying a current and
an upper state-of-charge breakpoint. The evaluator runs a PyBaMM
single-particle model with lumped thermal dynamics, SEI growth, and irreversible
lithium plating \cite{sulzer2021pybamm}. It simulates 30 cycles at each of
three ambient temperatures ($15$, $25$, and $35\,^{\circ}$C), including a
reference discharge for capacity measurement. A candidate is invalid if it
fails to reach $60\%$ state of charge within 600 seconds, exceeds
$65\,^{\circ}$C, crosses the plating-overpotential limit, violates the protocol
bounds, or fails numerically. For a valid protocol, an early-life capacity-fade
predictor \cite{severson2019battery} estimates the end-of-life cycle count;
the score is the three-temperature mean divided by 2,000.

\paragraph{ns-3 active queue management.}
The problem is to design an active queue management controller that maintains
high bottleneck utilization while limiting queueing delay, loss, and flow
imbalance \cite{almeida2024desired,toopchinezhad2025aqm}. The target program
implements \emph{control(obs)}: it observes queue length, average delay, recent
drops and ECN marks, recent utilization, and the number of active flows, then
returns RED/ECN thresholds, marking probability, queue limit, and an ECN
switch. A lightweight queue model calls the controller every $0.25$ seconds to
generate an action schedule, which is replayed in an ns-3 dumbbell simulation
\cite{henderson2008ns3}. The evaluator averages six 8-second packet-level
runs: long-flow, burst-dominated, and mixed traffic, each under two fixed seeds,
over a 20-Mb/s bottleneck. For each run, it computes
$g-d_{95}-2\ell-0.5(1-f)$, where $g$ is normalized goodput, $d_{95}$ is the
95th-percentile queueing delay in seconds, $\ell$ is loss rate, and $f$ is Jain
fairness; malformed actions or simulator failures invalidate the candidate.

\paragraph{FDTD Demux.}
The problem is to design a dielectric layout that routes two wavelengths from
a shared input waveguide to different output ports \cite{wu2024cwdm}. The
target program implements \emph{design(nx, ny, spec)} and returns a
resolution-parametric relative-permittivity pattern for the rectangular region
between one input and two output waveguides. The evaluator uses a fine-grid,
two-dimensional TM-mode FDTD simulation at both wavelengths. If $T_{iX}$
denotes normalized power from wavelength $i$ reaching output $X$, the score is
$0.5(T_{1A}+T_{2B})-0.25(T_{1B}+T_{2A})$: it rewards routing the first wavelength
to port A and the second to port B while penalizing crosstalk. The default real
grid is $480\times320$ cells with 10,000 time steps per wavelength. An incorrect
array shape, non-finite pattern or field, or failed simulation invalidates the
candidate.

\paragraph{Exothermic semi-batch reactor.}
The problem is to control a semi-batch reactor to maximize desired product
formation while avoiding thermal runaway and undesired side products
\cite{braniff2024risk}. Reactant B is fed into an initial charge of A; the
desired exothermic reaction $A+B\rightarrow C$ competes with the
temperature-accelerated side reaction $C\rightarrow D$. The target program
implements \emph{design(problem)} and returns ten piecewise-constant feed rates
and ten jacket temperatures for a one-hour batch. The evaluator integrates the
full stiff six-state material and energy balance under tight numerical
tolerances and averages $(n_C-n_D)/85$ over three operating scenarios with
different cooling capacities and runaway margins. Exceeding the
scenario-specific temperature limit or the 85-mol feed budget, producing a
malformed schedule or nonphysical state, or failing the solver invalidates the
candidate.

\paragraph{Perishable IRP.}
The problem is to choose daily replenishment quantities for perishable
inventory under uncertain demand while accounting for purchase, routing,
stockout, holding, and spoilage costs \cite{violi2024perishable}. The target
program implements \emph{policy(state)} and returns next-day shipment
quantities for eight retailers from age-bucket inventory, three-day forecasts,
recent demand, service levels, capacities, locations, and costs. The evaluator
enforces vehicle and store capacities, ages four-day-life inventory in FIFO
order, realizes candidate-independent demand, and constructs a canonical
nearest-neighbor delivery route. It runs 36 fixed independent 180-day episodes.
Each episode receives
$1-C/(12D)$, where $C$ is total purchase, holding, waste, lost-sales, vehicle,
and distance cost and $D$ is total demand; the final score is the episode mean
minus $0.15$ times its standard deviation. Invalid shipment vectors or capacity
violations invalidate the policy.

\section{Direct-LLM Baseline Protocol}
\label{sec:appendix-direct-llm}

\paragraph{Program-space adaptation of LAEA-Reg.}
LAEA uses an LLM as a surrogate for numerical decision vectors and provides a
regression-only variant, LAEA-Reg, in addition to its full regression and
classification pipeline \cite{hao2024llmsurrogate}. Our Direct-LLM
(LAEA-style) baseline retains the core regression mechanism but replaces a
fixed-dimensional vector with the source of a target program. We use this name
rather than LAEA because we do not transfer LAEA's numerical reproduction
operator, classification model, or unevaluated population to the free-form
program space.

\paragraph{Historical examples and prediction.}
At each promotion event, the baseline sorts completed real outcomes by
official utility and retains the best
\(\tau=\min(50,|\mathcal{H}|)\), matching the default history limit in the
released LAEA implementation. Real-valid utilities in this set are min--max
normalized to \([0,1]\), while invalid outcomes receive the common worst label
of zero. For each unevaluated candidate \(x\), the LLM receives the scientific
task and objective, the retained program sources and normalized real outcomes,
and the source of \(x\). One inference is made per candidate at temperature
zero. The response must contain only a scalar
\(\hat{y}_{\mathrm{LLM}}(x)\in[0,1]\) under the JSON key
\texttt{predicted\_quality}, either as a bare object or inside one JSON code
fence, with no additional fields or prose. Values outside this interval are
malformed. A
malformed response is retried twice; if all three attempts fail, the candidate
receives the worst prediction of zero.

\paragraph{Selection and real validation.}
Once the history contains at least two distinct real outcome labels, the
baseline promotes
\begin{equation}
x_{\mathrm{promote}} =
\arg\max_{x\in\mathcal{C}}\hat{y}_{\mathrm{LLM}}(x).
\label{eq:direct-llm-promotion}
\end{equation}
Thus, one valid and one invalid real outcome are sufficient to start
regression, whereas repeated invalid outcomes are not. Before this condition
holds, candidate generation follows the shared cold-start path and sends the
generated candidate directly to the real evaluator. The
baseline does not use a separate validity predictor, uncertainty, novelty,
adaptive gates, evaluator portfolios, regional credit, or periodic unranked
evaluation. Only a real-valid program may enter the algorithm population, and
every completed real outcome is added to the subsequent prediction history.

\section{Zero-Shot Reuse under Objective Reweighting}
\label{sec:appendix-zero-shot-reuse}

We test whether a component evaluator can be reused when objective weights
change, without collecting target labels. The FDTD Demux objective contains
four transmission terms: $T_{1A}$ and $T_{2B}$ measure the desired routing of
wavelength 1 to port A and wavelength 2 to port B, while $T_{1B}$ and $T_{2A}$
measure crosstalk into the incorrect ports. The original objective weights the
two desired terms equally at $(0.5,0.5)$; we change these weights to $(0.8,0.2)$
while keeping both crosstalk penalties at $-0.25$. We reuse the frozen,
source-calibrated component evaluator and apply the new weights directly to its
predictions of these four terms, without target labels or refitting. On 36
unseen candidates, the reweighted evaluator achieves an NDCG@3 of $0.959$ and
a normalized best regret@3 of $0.000$. Thus, its component representation
transfers immediately to the reweighted objective.

\section{LLM Backbone Sensitivity}
\label{sec:appendix-llm-backbone}

Table~\ref{tab:llm-backbone-sensitivity} evaluates backbone sensitivity on
PyBaMM at $B=70$. Gemini 3.1 Pro Preview yields higher AUBC$_{\Delta}$ and
terminal improvements for both search methods. With Gemini 2.5 Flash, Janus
has lower AUBC$_{\Delta}$ than Alg.-Only but achieves a higher terminal
improvement. This pattern suggests that the weaker backbone produces less
useful evaluator mutations and target-program proposals early in the search,
delaying improvements in proxy-guided budget allocation. As the real-outcome
archive grows, the LLM receives richer task feedback and more labels become
available for evaluator fitting and selection, allowing Janus to recover by
the end of the search.

\begin{table}[H]
\centering
\small
\setlength{\tabcolsep}{3.2pt}
\renewcommand{\arraystretch}{1.13}
\resizebox{\columnwidth}{!}{%
\begin{tabular}{@{}lcccccc@{}}
\toprule
\textbf{LLM Backbone} &
\multicolumn{2}{c}{\textbf{Janus}} &
\multicolumn{2}{c}{\textbf{Alg.-Only}} &
\multicolumn{2}{c}{\textbf{Advantage}} \\
\cmidrule(lr){2-3}\cmidrule(lr){4-5}\cmidrule(lr){6-7}
&
\shortstack{\textbf{AUBC$_{\Delta}$}\\\textbf{(\%)} $\uparrow$} &
\shortstack{\textbf{Best$_{\Delta}$@$B$}\\\textbf{(\%)} $\uparrow$} &
\shortstack{\textbf{AUBC$_{\Delta}$}\\\textbf{(\%)} $\uparrow$} &
\shortstack{\textbf{Best$_{\Delta}$@$B$}\\\textbf{(\%)} $\uparrow$} &
\shortstack{\textbf{AUBC$_{\Delta}$}\\\textbf{(pp)} $\uparrow$} &
\shortstack{\textbf{Best$_{\Delta}$@$B$}\\\textbf{(pp)} $\uparrow$} \\
\midrule
\textbf{Gemini 3.1 Pro Preview} & \textbf{312.9} & \textbf{362.6} & 274.2 & 288.3 & \textbf{$+38.7$} & \textbf{$+74.2$} \\
\textbf{Gemini 2.5 Flash} & 106.7 & \textbf{205.5} & \textbf{119.6} & 188.4 & $-12.9$ & \textbf{$+17.1$} \\
\bottomrule
\end{tabular}%
}
\caption{LLM-backbone sensitivity on PyBaMM at $B=70$.}
\label{tab:llm-backbone-sensitivity}
\end{table}

\section{Token-Cost Accounting}
\label{sec:appendix-token-cost}

We record the prompt and completion tokens used by each method and report the
average token count per completed real-evaluator call in
Table~\ref{tab:token-cost}.

\begin{table}[H]
\centering
\small
\setlength{\tabcolsep}{5.0pt}
\renewcommand{\arraystretch}{1.13}
\resizebox{\columnwidth}{!}{%
\begin{tabular}{@{}lrrrrr@{}}
\toprule
\textbf{Case} & \textbf{Janus} & \textbf{Alg.-Only} &
\textbf{Direct-LLM} & \textbf{Embed-GP} &
\shortstack{\textbf{Direct-LLM}\\\textbf{/ Janus}} \\
\midrule
\textbf{PyBaMM} & $20.1$ & $7.3$ & $119.2$ & $17.6$ & $5.9\times$ \\
\textbf{AQM} & $17.1$ & $5.1$ & $190.1$ & $17.7$ & $\mathbf{11.1\times}$ \\
\textbf{FDTD Demux} & $25.7$ & $5.7$ & $265.6$ & $21.6$ & $10.3\times$ \\
\textbf{Reactor} & $4.9$ & $2.7$ & $32.1$ & $14.4$ & $6.6\times$ \\
\textbf{Perishable IRP} & $7.9$ & $2.7$ & $53.5$ & $7.7$ & $6.8\times$ \\
\bottomrule
\end{tabular}%
}
\caption{Average LLM tokens per completed real-evaluator call
($10^3$ tokens; prompt plus completion; lower is better).}
\label{tab:token-cost}
\end{table}
\FloatBarrier

Janus uses $4.9$--$25.7$ thousand tokens per real-evaluator call across the five
cases. Direct-LLM consistently uses the most tokens, requiring
$5.9$--$11.1\times$ as many tokens as Janus, with a mean ratio of $8.1\times$.
Janus and Embed-GP have similar token counts in four cases: Janus is never more
than $1.2\times$ higher, while on Reactor it uses $2.9\times$ fewer tokens.
Alg.-Only has the lowest token count in every case; Janus uses
$1.8$--$4.5\times$ as many tokens as Alg.-Only.

\section{Evaluator Evolution Case Studies}
\label{sec:appendix-evaluator-cases}

The accuracy comparison in the main paper shows that evolved evaluators rank
held-out candidates more accurately than their seed versions. The three cases
below summarize the domain-relevant information introduced through evolution.

\paragraph{PyBaMM: degradation and safety features.}
The evolved evaluator augments aggregate charging outcomes with separate
indicators of peak temperature, lithium plating, SEI growth, voltage headroom,
and time spent at high state of charge. These signals help distinguish
protocols that appear similar in overall charging performance but differ in
degradation risk or proximity to safety constraints.

\paragraph{AQM: response and burst features.}
The evolved evaluator combines controller-structure information with controlled
response probes and burst-specific measurements. It captures how marking
probability and thresholds react to queue and utilization changes, while
retaining goodput, tail delay, and loss as separate signals under bursty
traffic. This gives the evaluator a more direct view of transient controller
behavior than a single aggregate rollout score.

\paragraph{FDTD: morphology features.}
The evolved evaluator supplements simulation scores with spatial descriptions
of the dielectric layout, including material density, variation, mirror
asymmetry, and top--bottom material imbalance. These morphology features help
differentiate layouts whose coarse simulations are similar but whose spatial
structure supports different routing behavior at the two output ports.

Across the three domains, evaluator evolution therefore enriches the proxy with
features tied to the real evaluator's decision criteria: physical safety and
degradation for charging, transient response for queue management, and spatial
morphology for photonic routing.

\endgroup

\end{document}